\pdfoutput=1
\documentclass[11pt]{article}

\usepackage[]{acl}

\usepackage{times}
\usepackage{latexsym}
\usepackage{amsmath}
\usepackage{amsfonts}
\usepackage{amssymb}
\usepackage{booktabs}
\usepackage{subcaption}
\usepackage{adjustbox}

\usepackage[T1]{fontenc}
\usepackage[utf8]{inputenc}

\usepackage{microtype}

\newcommand{\linear}{\textsc{Linear}}
\newcommand{\probeless}{\textsc{Probeless}}

\title{IDANI: Inference-time Domain Adaptation via Neuron-level Interventions}

\author{Omer Antverg \and Eyal Ben-David \and Yonatan Belinkov\thanks{~~Supported by the Viterbi Fellowship in the Center for Computer Engineering at the Technion.} \\
  Technion - Israel Institute of Technology \\
{\{omer.antverg@cs.$|$eyalbd12@campus.$|$belinkov@\}technion.ac.il}
}

\begin{document}
\maketitle
\begin{abstract}
Large pre-trained models are usually fine-tuned on downstream task data, and tested on unseen data.
When the train and test data come from different domains, the model is likely to struggle, as it is not adapted to the test domain.
% Based on a recent analysis of individual neurons in language models, 
We propose a new approach for domain adaptation (DA), using neuron-level interventions: 
We modify the representation of each test example in specific neurons,
% After fine-tuning the model on data from a source domain, given unlabeled samples from the test domain, we identify which neurons in the model encode information about the domain.
% We then modify the representation of each example only in these selected neurons, 
resulting in a counterfactual example from the source domain, which the model is more familiar with.
The modified example is then fed back into the model.
While most other DA methods are applied during training time, ours is applied during inference only, making it more efficient and applicable. 
Our experiments show that our method improves performance on unseen domains.
% As opposed to other DA methods, ours is applied during inference only.
\footnote{Our code is available at \url{https://github.com/technion-cs-nlp/idani}.} 
\end{abstract}

\vspace{-3pt}
\section{Introduction}
\vspace{-3pt}

A common assumption in NLP, and in machine learning in general, is that the training set and the test set are sampled from the same underlying distribution.
However, this assumption does not always hold in real-world applications since test data may arrive from many (target) domains, often not  seen during training.
Indeed, when applied to such unseen target domains, the trained model typically encounters significant degradation in performance.

DA algorithms aim to address this challenge by improving models' generalization to new domains, and algorithms for various DA scenarios have been developed \citep{daume2006domain, reichart-rappoport-2007-self, ben2007analysis, TACL183}. This work focuses on unsupervised domain adaptation (UDA), the most explored DA setup in recent years, which assumes access to labeled data from the source domain and unlabeled data from both source and target domains. Algorithms for this setup typically use the target domain knowledge during \emph{training}, attempting to bridge the gap between domains through representation learning \citep{DBLP:conf/acl/BlitzerDP07, JMLR:v17:15-239,ziser-reichart-2018-pivot,  DBLP:conf/emnlp/HanE19, TACL2255}.
Recently, \citet{bendavid2022pada} and \citet{volk2022example} introduced an approach for \emph{inference}-time DA, assuming no prior knowledge regarding the test domains but still modifying the training process to their gain.
% Despite the effectiveness of these approaches, the challenge of adaptation to target domains that were not seen is still underexplored.

% Domain adaptation (DA) algorithms aim to address this challenge by improving models' generalization to new domains.
% Most methods assume some access to target domain data during training~\citep[\textit{inter alia}]{daume2006domain, reichart-rappoport-2007-self, TACL183}, either labeled or unlabeled.
% Modern algorithms normally modify the model's fine-tuning procedure, attempting to make it learn domain-invariant representations, using pivot features~\citep{ziser-reichart-2018-pivot, TACL2255}, domain adversarial neural networks~\citep{JMLR:v17:15-239, DBLP:conf/iclr/0002ZWCMG18}, invariant risk minimization (IRM)~\citep{arjovsky2020invariant} or per-example prompts~\citep{bendavid2022pada}.

% We present an approach that performs UDA during inference time. By doing so, we overcome the need to train a new model for each new target domain. As a result, since test data may emanate from multiple domains, our algorithm is (much) more efficient than previous UDA approaches. Recently, \citet{bendavid2022pada} and \citet{volk2022example} introduced an approach for example-based adaptation during inference-time assuming no prior knowledge regarding the test domains.
In contrast to this line of work, we assume a more realistic scenario, in which the model was already trained on a source domain, and encounters unlabeled data from the target domain during inference time. 
%We assume access to the training data, but only use it during inference.
%Hence, our algorithm adapts to new domains on-the-fly.
%

\begin{figure*}[t]
    \vspace{-10pt}
    \centering
    \includegraphics[width=1\linewidth,keepaspectratio]{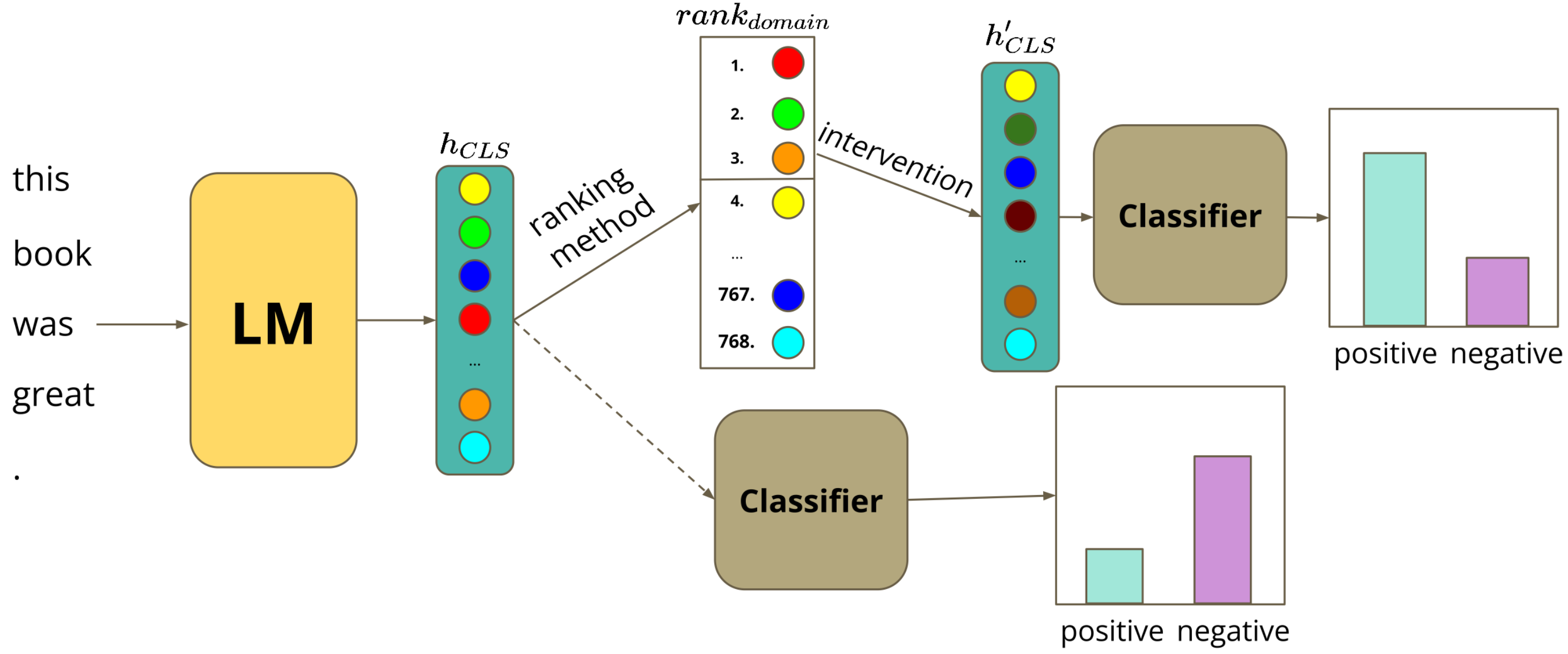}
    \caption{
    % Domain adaptation via individual neurons: 
    The language model---which was trained on some source domain, e.g., airline---creates a representation (CLS) for the review. Since the review is from a domain on which it was not trained, the model's classifier mistakenly classifies it as negative (bottom). In IDANI (top), the representation is fed into a neuron-ranking method.
    % , to rank its neurons by their importance for domain information. 
    The $k$-highest ranked neurons are modified by an intervention, to change the domain of the review, and the new representation is fed into the classifier, which correctly classifies it as positive.}
    \label{fig:da:illustration}
\end{figure*}

Given an example from a target domain, we would have liked to change it to a source domain example, so that the model would be more likely to perform well on it.
Since this is difficult to achieve, we aim to change its representation in a fine-grained manner, such that we modify only information about the domain of the representation, without hurting other information.
To do so, we take inspiration from work analyzing language models, which showed that linguistic properties are localized in certain neurons (dimensions in model representations) \citep{dalvi2019one,durrani-etal-2020-analyzing, torroba-hennigen-etal-2020-intrinsic, antverg2022pitfalls,Sajjad2021NeuronlevelIO}.
We first rank the neurons by their importance for identifying the domain (source or target) of each example. 
Then, we modify target-domain representations only in the highest-ranked neurons, to change their domain to the source domain.
Since the model was trained on examples from the source domain, we expect it to perform better on the modified representations.
We name this method as Inference-time Domain Adaptation via Neuron-level Interventions (IDANI). 

% Our method differs from classical DA algorithms in that it only requires an off-the-shelf fine-tuned model, without modifying the fine-tuning process and with no access to target domain samples at training time. 
% As such, it is easy and fast to employ.
% Also, our method requires only a single source domain, while other methods require multiple source domains or environments, as in IRM.

% In addition, this can be seen as an example for a real-world application of fine-grained control of language models.
% Previous work has shown the ability to control model's output~\citep{bau2018identifying, elazar2021amnesic, ravfogel-etal-2021-counterfactual, antverg2022pitfalls} from a theoretical point of view.
% In this work, we show a practical example for using this ability to improve model's performance on some tasks, through domain adaptation.
We follow a large body of previous work, testing IDANI on a variety of well known DA benchmarks, for a total of two text classification tasks (sentiment analysis, natural language inference) and one sequence tagging task (aspect identification), across 52 source--target domain pairs. We demonstrate that IDANI can improve results in many of these cases, with some significant gains.
% three tasks, and show that it can improve performance on unseen domains.
% \section{Related Work}
% \paragraph{Domain Adaptation}
% \paragraph{Individual Neurons}

\section{Method}

Given a model $M$ with a classification module $f$ and hidden dimensionality $d$, which was fine-tuned on data from a source domain $D_s=\{X_s\}$,
% we only assume access to labeled task data from a source domain $D_s=\{X_s,Y_s\}$, on which the model is fine-tuned with a classification module $f$. 
we receive unlabeled task data $D_t=\{X_t\}$ from a target domain for inference. 
As $s \neq t$, $M$'s performance is likely to deteriorate when processing $X_t$ compared to $X_s$.
Thus, we would like to make the representation of $X_t$ more similar to that of $X_s$ (regardless of the labels).
To do so, we apply the IDANI intervention method: 
\begin{enumerate}
    \item We process $X_s$ and $X_t$ through $M$, producing representations $H^s, H^t \subseteq \mathbb{R}^d$.
    We also compute $\bar{v}^s$ and $\bar{v}^t$, the element-wise mean representations of $X_s$ and $X_t$.
    \item We apply existing ranking methods to rank the representation's neurons by their relevance for domain information, i.e., the highest-ranked neuron holds the most information about the representation's domain (\S~\ref{da:rankings}).\footnote{Following previous work \citep{antverg2022pitfalls}, our method assumes that neurons with the same index carry similar information. While this is not necessarily true, we perform extrinsic (Table~\ref{tab:agg_results}) and intrinsic evaluations (Table~\ref{tab:da:examples}) that support this assumption.}
    \item For each $h^t \in H^t$, we would ideally like to have $h^s$, its source domain counterpart. Since $h^s$ is impossible to get, we create a counterfactual $\tilde{h}^s$ that simulates it by modifying $h^t$ only in the $k$-highest ranked neurons $\{n_1, ..., n_k\}$, such that $\forall i \in \{1 ,..., k\}$,
    \begin{equation}
        \tilde{h}^s_{n_i} = h^t_{n_i} + \alpha_{n_i} (\bar{v}^s_{n_i} - \bar{v}^t_{n_i})
    \end{equation}
    To allow stronger intervention on neurons that are ranked higher, we scale the intervention with 
 $\alpha \in \mathbb{R} ^d$,  a log-scaled sorted coefficients vector in the range $[0, \beta]$ such that $\alpha_{n_1}=\beta$ and $\alpha_{n_d}=0$, where $\beta$ is a hyperparameter~\citep{antverg2022pitfalls}.
    We denote the new set of representations as $\tilde{H}^s$.
    \item Representations from $\tilde{H}^s$ are fed into the classifier $f$---without re-training $f$---to predict the labels.
    Since $\tilde{H}^s$ is more similar to $H^s$ than $H^t$ is to $H^s$, we expect performance to improve.
    That is, for some scoring metric $\gamma$, we expect to have $\gamma(f(\tilde{H}^s)) > \gamma(f(H^t))$.
\end{enumerate}
The process is illustrated in Fig.~\ref{fig:da:illustration}.

\subsection{Ranking Methods}
\label{da:rankings}
We consider two ranking methods for ranking the representations' neurons (step 2):

\paragraph{\linear~\citep{dalvi2019one}}
This method trains a linear classifier on $H^s$ and $H^t$ to learn to predict the domain, using standard cross-entropy loss regularized by elastic net regularization~\citep{zou2005regularization}.
Then, it uses the classifier's weights to rank the neurons according to their importance for domain information.
Intuitively, neurons with a higher magnitude of absolute weights should be more important for predicting the domain.
% The choice of linear classifiers rather than deeper models is due to their explainability: the neuron-label weight naturally gives a measure of how indicative this neurons is to that specific label.
% With deeper models, it is unclear how to directly connect each neuron to the output.

\paragraph{\probeless}
The second ranking method is a simple one and does not rely on an external probe, and thus is very fast to obtain: it only depends on computing the mean representation of each domain ($\bar{v}^s$ and $\bar{v}^t)$, and sorting the difference between them.
% Formally, given $H^s$ and $H^t$, we compute $\bar{H}^s$ and $\bar{H}^t$, the element-wise mean representations of $H^s$ and $H^t$. 
For each neuron $i \in \{1,...,d\}$, we calculate the absolute difference between the means:
\begin{equation}
    r_i=|\bar{v}^s_i-\bar{v}^t_i|
\end{equation}
and obtain a ranking by arg-sorting $r$, i.e., the first neuron in the ranking corresponds to the highest value in $r$.
\citet{antverg2022pitfalls} showed that for interventions for morphology information, this method outperforms \linear\ and another ranking method~\citep{torroba-hennigen-etal-2020-intrinsic}.

\section{Experiments}
\subsection{Datasets}
We experiment with %three different datasets, which represent
%three different DA tasks, including 
two text classification tasks: sentiment analysis (classifying reviews to positive or negative~\citep{DBLP:conf/acl/BlitzerDP07}) and natural language inference (NLI; classifying whether two sentences entail or contradict each other~\citep{bowman2015large}), and a sequence tagging task: aspect prediction (identifying aspect terms within reviews~\citep{DBLP:conf/kdd/HuL04, DBLP:conf/acl/ToprakJG10, DBLP:conf/semeval/PontikiGPPAM14}).
For each task, the model is trained on a single source domain and tested on different target domains.
We explore a low-resource scenario, thus we use  2000--3000 examples from the source domain to  form  the training set.\footnote{For development data we split our training set in a ratio of 80:20, where the smaller portion is used for development. } For test, we use equivalent size data from the corresponding target domain.
% Overall we have 52 different source--target domain pairs.
Further data details are in Appendix~\ref{appendix:data}.

\subsection{Experiments}
For each task and pair of source and target domains, we fine-tune a pre-trained BERT-base-cased model~\citep{devlin-etal-2019-bert} on the training set of the source domain and evaluate its in-domain performance on the dev set of the source domain.\footnote{For all experimented models, we define a maximum sequence length value of 256 and use a training batch size of 16.}
We intervene on representations from the last layer of the model: word representations for the aspect prediction task, and CLS token representation for the other tasks.
We then test the model's out-of-distribution (OOD) performance on the test set of the target domain, for different $k$ (number of modified neurons) and $\beta$ (magnitude of the intervention) values: We perform grid search where $k$ is in the range $[0,d]$ ($d=768$) and $\beta$ is in the range $[1,10]$.
We experiment with both ranking methods described in \S~\ref{da:rankings}. 

We consider the model's performance at $k=0$ as its initial (unchanged) OOD performance (\textsc{init}), and report the difference between initial performance and performance using IDANI, with either \probeless\ ($\Delta^P$) or \linear\ ($\Delta^L$) rankings.
A limitation of IDANI (which we further discuss later) is the inability to choose the best $\beta$ and $k$ for each domain pair.
Following~\citet{antverg2022pitfalls} we report results for $\beta=8, k=50$ $(\Delta_{8,50})$, as well as oracle results (the best performance across all values, $\Delta_O$).
We consider the model's performance when fine-tuned on the target domain as an upper bound (\textsc{UB}). 
For all pairs, we repeat experiments using 5 different random seeds, and report mean \textsc{init}, $\Delta_{8,50}$, $\Delta_O$ and \textsc{UB} across seeds, alongside the standard error of the mean.

Since we assume that the model is exposed to target domain data only during inference, we cannot experiment with UDA methods, as they require access to the data during training. Furthermore, experimenting with \textit{inference-time DA} approaches \citep{bendavid2022pada, volk2022example} is also not possible since they assume multiple source domains for training.

\section{Results}
\label{da_results}
\begin{table}
\small
    \centering
    \begin{tabular}{c|cccc}
          & Improved & Damaged & Neither & AVG $\Delta$ \\ \midrule 
        $\Delta_{8,50}^P$ & $ 21 $ & $ 9 $ & $ 22 $ & $ 0.25 $\\ 
        $\Delta_{8,50}^L$ & $ 23 $ & $ 7 $ & $ 22 $ & $ 0.25 $\\ 
        % $\Delta_{S}^P$ & $ 9 $ & $ 1 $ & $ 42 $ \\ 
        % $\Delta_{S}^L$ & $ 9 $ & $ 1 $ & $ 42 $ \\ 
        $\Delta_{O}^P$ & $ 51 $ & $ 0 $ & $ 1 $  & $1.77$\\ 
        $\Delta_{O}^L$ & $ 50 $ & $ 0 $ & $ 2 $  & $0.93$ \\ 
    \bottomrule
    \end{tabular}
    \caption{Number of experiments in which IDANI improved, damaged, or did not significantly affect the initial performance. $\Delta^P$ and $\Delta^L$ refer to \probeless\ and \linear\ respectively, while $\Delta_{8,50}$ and $\Delta_{O}$ refer to $\beta=8, k=50$ and oracle values.}
    \label{tab:agg_results}
    \vspace{-5pt}
\end{table}
Overall, we have 52 source to target domain adaptation experiments.
Table~\ref{tab:agg_results} aggregates results across all experiments in three different categories: experiments where we can be confident that we improved the initial performance (i.e., the mean result across seeds is greater than the standard error), damaged it (mean lower than the negative standard error) or did not significantly affect it.
Detailed results per each source--target domain pair are in Appendix~\ref{appendix:da_res}.

% \subsection{\probeless\ Improves DA Results}
\begin{table*}
    \centering
    \begin{adjustbox}{width=\textwidth}
    \begin{tabular}{cp{10cm}}
    \toprule
         Airline  $\rightarrow$ DVD (Sentiment) & \textit{immortal, insanely, terrorist, crossing, obsessive, buzz, kidnapped} \\
         %frances, brutal, culminating}  \\ 
        %  Airline $\rightarrow$ Electronics (Sentiment Analysis) & \textit{strip, dedicated, laptops, placement, familiar, volt, longest, dsl, pushing, midrange} \\
         Laptops $\rightarrow$ Restaurant (Aspect) & \textit{Food, soup, selection, sushi, food, atmosphere, menu, staff} \\ 
         Restaurant $\rightarrow$ Laptops (Aspect) & \textit{time, user, slot, speed, MAC, Acer, system, size, SSD, design} \\
         \bottomrule
    \end{tabular}
    \end{adjustbox}
    \caption{Words that are part of sentences for which accuracy has improved the most (sentiment analysis), and words for which F1 score has improved the most (aspect prediction), using IDANI.}
    \label{tab:da:examples}
\end{table*}

As seen, IDANI provides decent performance, improving results much more than damaging even with default hyperparameters ($\Delta_{8,50}^P$ and $\Delta_{8,50}^L$).
With oracle hyperparameters ($\Delta_{O}^P$ and $\Delta_{O}^L$) it improves performance in almost all experiments.
% Even with non-oracle hyperparameters ($\Delta_{8,50}^P$ and $\Delta_{8,50}^L$), our method improves performance much more than damages it.

Some of these gains are quite impressive: In the aspect prediction task, we gain $18.8$ and $14.4$ F1 points when adapting the Restaurants source domain to the target domains Laptops and Service, respectively.
In other domain pairs, the gain is marginal.
On average we gain $4$ points with $\Delta_{O}^P$.

In sentiment analysis, the airline domain (A) is quite different from the others, leading to lower \textsc{init} (initial performance) scores when it is the source domain.
Adapting from A using IDANI results in a gain of up to $4.9$ accuracy points.
When other domains are used as source domains, we see mostly marginal gains, as the upper bound is closer to the initial performance, leaving less room for improvement in this task ($\textsc{ub}  - \textsc{init}$ is low).

In NLI, it seems harder to improve: the room for improvement is lower ($3.3$ F1 points on average), which may imply that domain information is not crucial for this task. 
Still, we do see some significant gains, e.g., an improvement of $2$ F1 points when adapting from Slate to the Telephone domain.

Generally, across all tasks and domain pairs, \probeless\ provides better performance than \linear\, as $\Delta_{O}^P > \Delta_{O}^L$ in 47 of the 52 experiments (Appendix~\ref{appendix:da_res}). 
This is in line with the insights from~\citet{antverg2022pitfalls}, who observed that \probeless\ was better than \linear\ when used for intervening on morphological attributes.

\subsection{Qualitative Analysis}
To analyze the benefits of IDANI, for each word in the dataset we record the change in results when classifying sentences containing the word (sentiment analysis) or when classifying the word itself (aspect prediction). 
We report the words with the greatest improvement in Table~\ref{tab:da:examples}.
When switching from the Airline domain to the DVD domain in the sentiment analysis task, those are mostly words that sound negative in an airline context, but may not imply a sentiment towards a movie (\textit{terrorist}, \textit{kidnapped}).
In the aspect prediction task, those are mostly target domain related terms that are not likely to appear in the source domain.

\subsection{Default $\beta$ and $k$ are Not Optimal}
\label{da:limitation}
\begin{figure}
\vspace{-5pt}
    \centering
    \includegraphics[width=1\columnwidth]{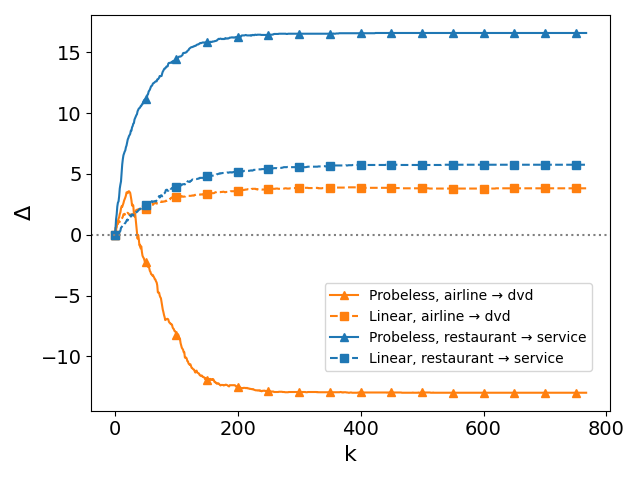}
    \caption{Results for different $k$ values, using $\beta=8$.}
    \label{fig:k}
\end{figure}
% \begin{figure*}
%     \begin{subfigure}[t]{0.48\textwidth}
%         \centering
%         \includegraphics[width=1\linewidth]{figures/airline_dvd_beta_8.png}
%         \caption{$\beta=8$.}
%         \label{fig:airline_dvd_beta}    
%     \end{subfigure}
%     \begin{subfigure}[t]{0.48\textwidth}
%         \centering
%         \includegraphics[width=1\linewidth]{figures/airline_dvd_k_50.png}
%         \caption{$k=50$.}
%         \label{fig:airline_dvd_k}
%     \end{subfigure}
%     \caption{Sentiment analysis, Airline to DVD results for different $k$ values with $\beta=8$ (left) and for different $\beta$ values with $k=50$ (right).}
%     \label{fig:airline_dvd}
% \end{figure*}

% \begin{figure*}
%     \begin{subfigure}[t]{0.48\textwidth}
%         \centering
%         \includegraphics[width=1\linewidth]{figures/rest_service_beta_8.png}
%         \caption{$\beta=8$.}
%         \label{fig:rest_service_beta}    
%     \end{subfigure}
%     \begin{subfigure}[t]{0.48\textwidth}
%         \centering
%         \includegraphics[width=1\linewidth]{figures/rest_service_k_50.png}
%         \caption{$k=50$.}
%         \label{fig:rest_service_k}
%     \end{subfigure}
%     \caption{Aspect prediction, Restaurant to Service results for different $k$ values with $\beta=8$ (left) and for different $\beta$ values with $k=50$ (right).}
%     \label{fig:rest_service}
% \end{figure*}
While the potential for performance improvement with \probeless\ is high, the selection of $\beta=8, k=50$ turns out as non optimal, as $\Delta_{8,50}^P$ is well below $\Delta_{O}^P$ across our experiments.
This is also true for $\Delta_{8,50}^L$ compared to $\Delta_{O}^L$, but to a lesser degree.

% There are cases where the default hyperparameters values damage performance.
% One example is when the source domain is Airline in the sentiment analysis task.
% : while $\Delta_{O}^P$ gains at least $2.3$ accuracy points adapting to any of the target domains, choosing $\beta=8, k=50$ hurts performance by up to $4.4$ points. 
Fig.~\ref{fig:k} shows that a milder intervention---lower $k$ value---would have been more ideal for the Airline $\rightarrow$ DVD scenario. 
Modifying too many neurons probably affects other encoded information---besides domain information---damaging the task performance.
Thus, we might lean towards smaller $k$ values.
However, this is not always the case: Fig.~\ref{fig:k} also shows that for the Restaurant $\rightarrow$ Service scenario in the aspect prediction task, \probeless' performance reaches a saturation point around the value of $k=100$ neurons. Thus there is no ideal value of $k$ across all domain pairs.
A similar phenomenon with $\beta$ is shown in Appendix~\ref{appendix:hyper}.

% It thus seems that there are no values that would be optimal for these parameters across all domain pairs.
Therefore, hyperparameters should be task- and domain-dependent, but it is unclear how to define them for each domain pair.
% Therefore, for each pair of domains, we would like to select the values that best fit it.
% However, with no access to any labeled data from the target domain, this is not an easy task: preliminary experiments to select $\beta$ and $k$ based on heuristics such as the distance between the domains' means, or clustering methods, failed to provide good results.
Yet, 
% we should note that
% the restriction which we assume in this work---no labeled target domain data---is a harsh one, and 
in most real-world cases some labeled data should be available or could be manually created.
In such cases, the best approach would be to grid-search over the hyperparameters on the available labeled data, and use the selected values for the (unlabeled) test data.

\section{Conclusion}
In this work, we demonstrated the ability to leverage neuron-intervention methods to improve OOD performance.
We showed that in some cases, IDANI can significantly help models to adapt to new domains.
% In consistence with~\citet{antverg2022pitfalls}, \probeless\ provides better results than \linear, while also being faster to apply.
IDANI performs best with oracle hyperparameters, but even with the default ones we see overall positive results.
We showed that IDANI indeed focuses on domain-related information, as the gains come mostly from domain-related information, such as domain-specific aspect terms.
Importantly, IDANI is applied only during inference, unlike most other DA methods.

\section*{Acknowledgements}
This research was supported by the ISRAEL SCIENCE FOUNDATION (grant No.\ 448/20) and by an Azrieli Foundation Early Career Faculty Fellowship. We also thank the anonymous reviewers for their insightful comments and suggestions.

\bibliography{anthology,custom}

\begin{thebibliography}{25}
\expandafter\ifx\csname natexlab\endcsname\relax\def\natexlab#1{#1}\fi

\bibitem[{Antverg and Belinkov(2022)}]{antverg2022pitfalls}
Omer Antverg and Yonatan Belinkov. 2022.
\newblock \href {https://openreview.net/forum?id=8uz0EWPQIMu} {On the pitfalls
  of analyzing individual neurons in language models}.
\newblock In \emph{International Conference on Learning Representations}.

\bibitem[{Ben-David et~al.(2021)Ben-David, Oved, and
  Reichart}]{bendavid2022pada}
Eyal Ben-David, Nadav Oved, and Roi Reichart. 2021.
\newblock Pada: A prompt-based autoregressive approach for adaptation to unseen
  domains.
\newblock \emph{arXiv preprint arXiv:2102.12206}.

\bibitem[{Ben-David et~al.(2007)Ben-David, Blitzer, Crammer, Pereira
  et~al.}]{ben2007analysis}
Shai Ben-David, John Blitzer, Koby Crammer, Fernando Pereira, et~al. 2007.
\newblock Analysis of representations for domain adaptation.
\newblock \emph{Advances in neural information processing systems}, 19:137.

\bibitem[{Blitzer et~al.(2007)Blitzer, Dredze, and
  Pereira}]{DBLP:conf/acl/BlitzerDP07}
John Blitzer, Mark Dredze, and Fernando Pereira. 2007.
\newblock \href {https://www.aclweb.org/anthology/P07-1056/} {Biographies,
  bollywood, boom-boxes and blenders: Domain adaptation for sentiment
  classification}.
\newblock In \emph{{ACL} 2007, Proceedings of the 45th Annual Meeting of the
  Association for Computational Linguistics, June 23-30, 2007, Prague, Czech
  Republic}. The Association for Computational Linguistics.

\bibitem[{Bowman et~al.(2015)Bowman, Angeli, Potts, and
  Manning}]{bowman2015large}
Samuel~R. Bowman, Gabor Angeli, Christopher Potts, and Christopher~D. Manning.
  2015.
\newblock \href {https://doi.org/10.18653/v1/D15-1075} {A large annotated
  corpus for learning natural language inference}.
\newblock In \emph{Proceedings of the 2015 Conference on Empirical Methods in
  Natural Language Processing}, pages 632--642, Lisbon, Portugal. Association
  for Computational Linguistics.

\bibitem[{Dalvi et~al.(2019)Dalvi, Durrani, Sajjad, Belinkov, Bau, and
  Glass}]{dalvi2019one}
Fahim Dalvi, Nadir Durrani, Hassan Sajjad, Yonatan Belinkov, Anthony Bau, and
  James~R. Glass. 2019.
\newblock \href {https://doi.org/10.1609/aaai.v33i01.33016309} {What is one
  grain of sand in the desert? analyzing individual neurons in deep {NLP}
  models}.
\newblock In \emph{The Thirty-Third {AAAI} Conference on Artificial
  Intelligence, {AAAI} 2019, The Thirty-First Innovative Applications of
  Artificial Intelligence Conference, {IAAI} 2019, The Ninth {AAAI} Symposium
  on Educational Advances in Artificial Intelligence, {EAAI} 2019, Honolulu,
  Hawaii, USA, January 27 - February 1, 2019}, pages 6309--6317. {AAAI} Press.

\bibitem[{Daume~III and Marcu(2006)}]{daume2006domain}
Hal Daume~III and Daniel Marcu. 2006.
\newblock Domain adaptation for statistical classifiers.
\newblock \emph{Journal of artificial Intelligence research}, 26:101--126.

\bibitem[{David et~al.(2020)David, Rabinovitz, and Reichart}]{TACL2255}
Eyal~Ben David, Carmel Rabinovitz, and Roi Reichart. 2020.
\newblock \href {https://transacl.org/ojs/index.php/tacl/article/view/2255}
  {Perl: Pivot-based domain adaptation for pre-trained deep contextualized
  embedding models}.
\newblock \emph{Transactions of the Association for Computational Linguistics},
  8(0):504--521.

\bibitem[{Devlin et~al.(2019)Devlin, Chang, Lee, and
  Toutanova}]{devlin-etal-2019-bert}
Jacob Devlin, Ming-Wei Chang, Kenton Lee, and Kristina Toutanova. 2019.
\newblock \href {https://doi.org/10.18653/v1/N19-1423} {{BERT}: Pre-training of
  deep bidirectional transformers for language understanding}.
\newblock In \emph{Proceedings of the 2019 Conference of the North {A}merican
  Chapter of the Association for Computational Linguistics: Human Language
  Technologies, Volume 1 (Long and Short Papers)}, pages 4171--4186,
  Minneapolis, Minnesota. Association for Computational Linguistics.

\bibitem[{Durrani et~al.(2020)Durrani, Sajjad, Dalvi, and
  Belinkov}]{durrani-etal-2020-analyzing}
Nadir Durrani, Hassan Sajjad, Fahim Dalvi, and Yonatan Belinkov. 2020.
\newblock \href {https://doi.org/10.18653/v1/2020.emnlp-main.395} {Analyzing
  individual neurons in pre-trained language models}.
\newblock In \emph{Proceedings of the 2020 Conference on Empirical Methods in
  Natural Language Processing (EMNLP)}, pages 4865--4880, Online. Association
  for Computational Linguistics.

\bibitem[{Ganin et~al.(2016)Ganin, Ustinova, Ajakan, Germain, Larochelle,
  Laviolette, March, and Lempitsky}]{JMLR:v17:15-239}
Yaroslav Ganin, Evgeniya Ustinova, Hana Ajakan, Pascal Germain, Hugo
  Larochelle, Fran{\c{c}}ois Laviolette, Mario March, and Victor Lempitsky.
  2016.
\newblock \href {http://jmlr.org/papers/v17/15-239.html} {Domain-adversarial
  training of neural networks}.
\newblock \emph{Journal of Machine Learning Research}, 17(59):1--35.

\bibitem[{Gong et~al.(2020)Gong, Yu, and Xia}]{DBLP:conf/emnlp/GongYX20}
Chenggong Gong, Jianfei Yu, and Rui Xia. 2020.
\newblock \href {https://doi.org/10.18653/v1/2020.emnlp-main.572} {Unified
  feature and instance based domain adaptation for aspect-based sentiment
  analysis}.
\newblock In \emph{Proceedings of the 2020 Conference on Empirical Methods in
  Natural Language Processing, {EMNLP} 2020, Online, November 16-20, 2020},
  pages 7035--7045. Association for Computational Linguistics.

\bibitem[{Han and Eisenstein(2019)}]{DBLP:conf/emnlp/HanE19}
Xiaochuang Han and Jacob Eisenstein. 2019.
\newblock \href {https://doi.org/10.18653/v1/D19-1433} {Unsupervised domain
  adaptation of contextualized embeddings for sequence labeling}.
\newblock In \emph{Proceedings of the 2019 Conference on Empirical Methods in
  Natural Language Processing and the 9th International Joint Conference on
  Natural Language Processing, {EMNLP-IJCNLP} 2019, Hong Kong, China, November
  3-7, 2019}, pages 4237--4247. Association for Computational Linguistics.

\bibitem[{Hu and Liu(2004)}]{DBLP:conf/kdd/HuL04}
Minqing Hu and Bing Liu. 2004.
\newblock \href {https://doi.org/10.1145/1014052.1014073} {Mining and
  summarizing customer reviews}.
\newblock In \emph{Proceedings of the Tenth {ACM} {SIGKDD} International
  Conference on Knowledge Discovery and Data Mining, Seattle, Washington, USA,
  August 22-25, 2004}, pages 168--177. {ACM}.

\bibitem[{Nguyen(2015)}]{Nguyen2015airline}
Quang Nguyen. 2015.
\newblock \href {https://github.com/quankiquanki/skytrax-reviews-dataset} {The
  airline review dataset}.

\bibitem[{Pontiki et~al.(2014)Pontiki, Galanis, Pavlopoulos, Papageorgiou,
  Androutsopoulos, and Manandhar}]{DBLP:conf/semeval/PontikiGPPAM14}
Maria Pontiki, Dimitris Galanis, John Pavlopoulos, Harris Papageorgiou, Ion
  Androutsopoulos, and Suresh Manandhar. 2014.
\newblock \href {https://doi.org/10.3115/v1/s14-2004} {Semeval-2014 task 4:
  Aspect based sentiment analysis}.
\newblock In \emph{Proceedings of the 8th International Workshop on Semantic
  Evaluation, SemEval@COLING 2014, Dublin, Ireland, August 23-24, 2014}, pages
  27--35. The Association for Computer Linguistics.

\bibitem[{Reichart and Rappoport(2007)}]{reichart-rappoport-2007-self}
Roi Reichart and Ari Rappoport. 2007.
\newblock \href {https://aclanthology.org/P07-1078} {Self-training for
  enhancement and domain adaptation of statistical parsers trained on small
  datasets}.
\newblock In \emph{Proceedings of the 45th Annual Meeting of the Association of
  Computational Linguistics}, pages 616--623, Prague, Czech Republic.
  Association for Computational Linguistics.

\bibitem[{Sajjad et~al.(2021)Sajjad, Durrani, and
  Dalvi}]{Sajjad2021NeuronlevelIO}
Hassan Sajjad, Nadir Durrani, and Fahim Dalvi. 2021.
\newblock Neuron-level interpretation of deep nlp models: A survey.
\newblock \emph{ArXiv}, abs/2108.13138.

\bibitem[{Schnabel and Schütze(2014)}]{TACL183}
Tobias Schnabel and Hinrich Schütze. 2014.
\newblock \href {https://transacl.org/ojs/index.php/tacl/article/view/183}
  {Flors: Fast and simple domain adaptation for part-of-speech tagging}.
\newblock \emph{Transactions of the Association for Computational Linguistics},
  2(0):15--26.

\bibitem[{Toprak et~al.(2010)Toprak, Jakob, and
  Gurevych}]{DBLP:conf/acl/ToprakJG10}
Cigdem Toprak, Niklas Jakob, and Iryna Gurevych. 2010.
\newblock \href {https://www.aclweb.org/anthology/P10-1059/} {Sentence and
  expression level annotation of opinions in user-generated discourse}.
\newblock In \emph{{ACL} 2010, Proceedings of the 48th Annual Meeting of the
  Association for Computational Linguistics, July 11-16, 2010, Uppsala,
  Sweden}, pages 575--584. The Association for Computer Linguistics.

\bibitem[{Torroba~Hennigen et~al.(2020)Torroba~Hennigen, Williams, and
  Cotterell}]{torroba-hennigen-etal-2020-intrinsic}
Lucas Torroba~Hennigen, Adina Williams, and Ryan Cotterell. 2020.
\newblock \href {https://doi.org/10.18653/v1/2020.emnlp-main.15} {Intrinsic
  probing through dimension selection}.
\newblock In \emph{Proceedings of the 2020 Conference on Empirical Methods in
  Natural Language Processing (EMNLP)}, pages 197--216, Online. Association for
  Computational Linguistics.

\bibitem[{Volk et~al.(2022)Volk, Ben-David, Amosy, Chechik, and
  Reichart}]{volk2022example}
Tomer Volk, Eyal Ben-David, Ohad Amosy, Gal Chechik, and Roi Reichart. 2022.
\newblock Example-based hypernetworks for out-of-distribution generalization.
\newblock \emph{arXiv preprint arXiv:2203.14276}.

\bibitem[{Williams et~al.(2018)Williams, Nangia, and
  Bowman}]{williams2018broad}
Adina Williams, Nikita Nangia, and Samuel~R Bowman. 2018.
\newblock \href {https://www.aclweb.org/anthology/N18-1101/} {A broad-coverage
  challenge corpus for sentence understanding through inference}.
\newblock In \emph{2018 Conference of the North American Chapter of the
  Association for Computational Linguistics: Human Language Technologies, NAACL
  HLT 2018}, pages 1112--1122.

\bibitem[{Ziser and Reichart(2018)}]{ziser-reichart-2018-pivot}
Yftah Ziser and Roi Reichart. 2018.
\newblock \href {https://doi.org/10.18653/v1/N18-1112} {Pivot based language
  modeling for improved neural domain adaptation}.
\newblock In \emph{Proceedings of the 2018 Conference of the North {A}merican
  Chapter of the Association for Computational Linguistics: Human Language
  Technologies, Volume 1 (Long Papers)}, pages 1241--1251, New Orleans,
  Louisiana. Association for Computational Linguistics.

\bibitem[{Zou and Hastie(2005)}]{zou2005regularization}
Hui Zou and Trevor Hastie. 2005.
\newblock Regularization and variable selection via the elastic net.
\newblock \emph{Journal of the royal statistical society: series B (statistical
  methodology)}, 67(2):301--320.

\end{thebibliography}
\bibliographystyle{acl_natbib}

\newpage
\appendix

\section{Data Details}
\label{appendix:data}
We test IDANI on three different tasks: sentiment analysis, natural language inference, and aspect prediction. Further details of the training, development, and test sets of each domain are provided in Table~\ref{tab:data-stats}.

\paragraph{Sentiment Analysis}
We follow a large body of prior DA work to focus on the task of binary sentiment classification. 
We experiment with the four legacy product review domains of~\citet{DBLP:conf/acl/BlitzerDP07}: Books (B), DVDs (D), Electronic items (E) and Kitchen appliances (K). 
We also experiment in a more challenging setup, considering an airline review dataset (A)~\citep{Nguyen2015airline, ziser-reichart-2018-pivot}. 
This setup is more challenging because of the differences between the product and service domains. 

\paragraph{Natural Language Inference}
\citep{williams2018broad}
This corpus is an extension of the SNLI dataset \citep{bowman2015large}. 
Each example consists of a pair of sentences, a premise and a hypothesis. 
The relationship between the two may be entailment, contradiction, or neutral. 
The corpus includes data from $10$ domains: $5$ are matched, with training, development and test sets, and $5$ are mismatched, without a training set. 
Following~\citet{bendavid2022pada}, we experiment only with the five matched domains: Fiction (F), Government (G), Slate (SL), Telephone (TL) and Travel (TR).

Since the test sets of the MNLI dataset are not publicly available, we use the original development sets as our test sets for each target domain, while source domains use these sets for development. 
Following prior work~\citep{bendavid2022pada, volk2022example} we explore a low-resource supervised scenario, which emphasizes the need for a DA algorithm. 
Thus, we randomly downsample each of the training sets by a factor of $30$, resulting in 2,000--3,000 examples per set.

\paragraph{Aspect Prediction}
The aspect prediction dataset is based on  aspect-based sentiment analysis (ABSA) corpora from four domains: Device (D), Laptops (L), Restaurant (R), and Service (SE). 
The D data consists of reviews from \citet{DBLP:conf/acl/ToprakJG10}, the SE data includes web service reviews \citep{DBLP:conf/kdd/HuL04}, and the L and R domains consist of reviews from the SemEval-2014 ABSA challenge \citep{DBLP:conf/semeval/PontikiGPPAM14}.
The task is to identify aspect terms within reviews.
For example, given a sentence ``The price is reasonable, although the service is poor'', both ``price'' and ``service'' should be identified as aspect terms.

We follow the training and test splits defined by \citet{DBLP:conf/emnlp/GongYX20} for the D and SE domains, while the splits for the L and R domains are taken from the SemEval-2014 ABSA challenge. 
To establish our development set, we randomly sample $10\%$ out of the training data. 

\begin{table}
\small
\centering
\begin{adjustbox}{width=0.47\textwidth}
\begin{tabular}{l c c c }
\toprule
    \multicolumn{4}{c}{\textbf{Sentiment Classification}} \\
  \midrule
  \textbf{Domain} & \textbf{Training (src)}  & \textbf{Dev (src)} & \textbf{Test (trg)} \\
   
  \midrule
  \textbf{Airline (A)} & $1,700$ & $300$ & $2,000$ \\
  \textbf{Books (B)} & $1,700$ & $300$ & $2,000$ \\
  \textbf{DVD (D)} & $1,700$ & $300$ & $2,000$ \\
  \textbf{Electronics (E)} & $1,700$ & $300$ & $2,000$ \\
  \textbf{Kitchen (K)} & $1,700$ & $300$ & $2,000$ \\
  \midrule
  \multicolumn{4}{c}{\textbf{MNLI} }\\
  \midrule
  \textbf{Domain} & \textbf{Training (src)}  & \textbf{Dev (src)} & \textbf{Test (trg)} \\
   
  \midrule
  \textbf{Fiction (F)} & $2,547$ & $1,972$ & $1,972$ \\
  \textbf{Government (G)} & $2,541$ & $1,944$ & $1,944$ \\
  \textbf{Slate (SL)} & $2,605$ & $1,954$ & $1,954$ \\
  \textbf{Telephone(TL)} & $2,754$ & $1,965$ & $1,965$ \\
  \textbf{Travel (TR)} & $2,541$ & $1,975$ & $1,975$ \\
  \midrule
  \multicolumn{4}{c}{\textbf{Aspect} }\\
  \midrule
  \textbf{Domain} & \textbf{Training (src)}  & \textbf{Dev (src)} & \textbf{Test (trg)} \\
   
  \midrule
  \textbf{Device (D)} & $2,302$ & $255$ & $1,279$ \\
  \textbf{Laptops (L)} & $2,726$ & $303$ & $800$ \\
  \textbf{Restaurants (R)} & $3,487$ & $388$ & $800$ \\
  \textbf{Service(SE)} & $1,343$ & $149$ & $747$ \\
  \bottomrule
\end{tabular}
\end{adjustbox}
\caption{The number of examples in each domain of our four tasks. We denote the examples used when a domain is the source domain (src), and when it is the target domain (trg).}
\label{tab:data-stats}
\end{table}

\section{Detailed Results}
\label{appendix:da_res}
Results for all domain pairs are shown in Tables~\ref{tab:res:app:sent},~\ref{tab:res:app:mnli} and~\ref{tab:res:app:aspect}.
As described in \S~\ref{da_results}, IDANI can potentially significantly improve performance, shown by the results of $\Delta_{O}^P$.
Current hyperparameter values do not fulfill this entire potential, but still improve performance in most cases ($\Delta_{8,50}^P$).

\begin{table*}
\small
\begin{subtable}{1\textwidth}
\centering
\begin{adjustbox}{width=\textwidth}
\begin{tabular}{cccccccc} \toprule
 & A $\rightarrow$ B & A $\rightarrow$ D & A $\rightarrow$ E & A $\rightarrow$ K & B $\rightarrow$ A & B $\rightarrow$ D & B $\rightarrow$ E \\ 
 \midrule
  \textsc{init} & $ 77.4 \pm 1.3 $ & $ 75.5 \pm 2.2 $ & $ 85.2 \pm 1.0 $ & $ 84.9 \pm 0.9 $ & $ 83.7 \pm 0.7 $ & $ 87.9 \pm 0.3 $ & $ 90.4 \pm 0.2 $ \\
  $\textsc{UB}$ & $ 88.0 \pm 0.5 $ & $ 89.2 \pm 0.5 $ & $ 92.4 \pm 0.4 $ & $ 92.4 \pm 0.2 $ & $ 88.0 \pm 0.1 $ & $ 89.2 \pm 0.5 $ & $ 92.4 \pm 0.4 $ \\ 
  $\Delta_{8,50}^P$ & $ -4.4 \pm 4.8 $ & $ -2.2 \pm 5.4 $ & $ -1.2 \pm 2.4 $ & $ -1.5 \pm 1.9 $ & $ 0.5 \pm 0.1 $ & $ 0.1 \pm 0.1 $ & $ -0.0 \pm 0.0 $ \\ 
  $\Delta_{8,50}^L$ & $ 2.0 \pm 1.0 $ & $ 2.1 \pm 1.0 $ & $ 1.3 \pm 0.4 $ & $ 1.1 \pm 0.5 $ & $ 0.2 \pm 0.1 $ & $ 0.1 \pm 0.0 $ & $ -0.0 \pm 0.0 $ \\ 
  % $\Delta_{S}^P$ & $ 2.6 \pm 1.3 $ & $ 3.9 \pm 1.6 $ & $ 1.7 \pm 0.8 $ & $ 0.9 \pm 0.4 $ & $ 0.2 \pm 0.1 $ & $ 0.0 \pm 0.0 $ & $ 0.0 \pm 0.0 $ \\
  % $\Delta_{S}^L$ & $ 2.5 \pm 1.3 $ & $ 3.7 \pm 1.5 $ & $ 1.8 \pm 0.7 $ & $ 0.9 \pm 0.4 $ & $ 0.0 \pm 0.0 $ & $ 0.0 \pm 0.0 $ & $ 0.0 \pm 0.0 $ \\
  $\Delta_{O}^P$ & $ 3.0 \pm 1.3 $ & $ 4.9 \pm 1.8 $ & $ 2.3 \pm 0.8 $ & $ 2.3 \pm 1.0 $ & $ 0.9 \pm 0.2 $ & $ 0.3 \pm 0.1 $ & $ 0.1 \pm 0.0 $ \\ 
  $\Delta_{O}^L$ & $ 2.9 \pm 1.3 $ & $ 4.2 \pm 1.8 $ & $ 2.3 \pm 0.8 $ & $ 2.2 \pm 0.9 $ & $ 0.3 \pm 0.1 $ & $ 0.1 \pm 0.0 $ & $ 0.0 \pm 0.0 $ \\
\end{tabular}
\end{adjustbox}
\end{subtable}
\begin{subtable}{1\textwidth}
\centering
\begin{adjustbox}{width=\textwidth}
\begin{tabular}{cccccccc} \midrule
 & B $\rightarrow$ K & D $\rightarrow$ A & D $\rightarrow$ B & D $\rightarrow$ E & D $\rightarrow$ K & E $\rightarrow$ A & E $\rightarrow$ B \\ 
 \midrule
  \textsc{init} & $ 87.8 \pm 0.4 $ & $ 81.5 \pm 0.3 $ & $ 89.4 \pm 0.3 $ & $ 90.3 \pm 0.2 $ & $ 88.1 \pm 0.5 $ & $ 86.3 \pm 0.4 $ & $ 86.8 \pm 0.4 $ \\ 
  $\textsc{UB}$ & $ 92.4 \pm 0.2 $ & $ 88.0 \pm 0.1 $ & $ 88.0 \pm 0.5 $ & $ 92.4 \pm 0.4 $ & $ 92.4 \pm 0.2 $ & $ 88.0 \pm 0.1 $ & $ 88.0 \pm 0.5 $ \\
  $\Delta_{8,50}^P$ & $ 0.1 \pm 0.0 $ & $ 0.8 \pm 0.2 $ & $ 0.1 \pm 0.1 $ & $ -0.0 \pm 0.1 $ & $ 0.8 \pm 0.3 $ & $ 0.0 \pm 0.0 $ & $ 0.6 \pm 0.2 $ \\ 
  $\Delta_{8,50}^L$ & $ 0.1 \pm 0.0 $ & $ 0.5 \pm 0.1 $ & $ 0.1 \pm 0.0 $ & $ 0.1 \pm 0.0 $ & $ 0.2 \pm 0.1 $ & $ 0.0 \pm 0.0 $ & $ 0.1 \pm 0.1 $ \\
  % $\Delta_{S}^P$ & $ 0.0 \pm 0.0 $ & $ 1.0 \pm 0.4 $ & $ 0.2 \pm 0.1 $ & $ 0.0 \pm 0.0 $ & $ 0.3 \pm 0.2 $ & $ 0.0 \pm 0.0 $ & $ 0.7 \pm 0.3 $ \\
  % $\Delta_{S}^L$ & $ -0.0 \pm 0.0 $ & $ 0.4 \pm 0.2 $ & $ 0.1 \pm 0.1 $ & $ 0.0 \pm 0.0 $ & $ 0.2 \pm 0.2 $ & $ 0.0 \pm 0.0 $ & $ 0.1 \pm 0.1 $ \\ 
  $\Delta_{O}^P$ & $ 0.4 \pm 0.1 $ & $ 1.4 \pm 0.3 $ & $ 0.3 \pm 0.1 $ & $ 0.3 \pm 0.1 $ & $ 1.4 \pm 0.5 $ & $ 0.2 \pm 0.0 $ & $ 1.0 \pm 0.3 $ \\ 
  $\Delta_{O}^L$ & $ 0.2 \pm 0.0 $ & $ 0.8 \pm 0.1 $ & $ 0.2 \pm 0.1 $ & $ 0.1 \pm 0.0 $ & $ 0.5 \pm 0.2 $ & $ 0.1 \pm 0.0 $ & $ 0.3 \pm 0.1 $ \\
%   \bottomrule
\end{tabular}
\end{adjustbox}
\end{subtable}
\begin{subtable}{1\textwidth}
\centering
\begin{adjustbox}{width=\textwidth}
\begin{tabular}{ccccccc|c} \midrule
 & E $\rightarrow$ D & E $\rightarrow$ K & K $\rightarrow$ A & K $\rightarrow$ B & K $\rightarrow$ D & K $\rightarrow$ E & AVG \\ 
 \midrule
  \textsc{init} & $ 86.5 \pm 0.2 $ & $ 93.2 \pm 0.3 $ & $ 83.9 \pm 0.4 $ & $ 87.0 \pm 0.2 $ & $ 86.4 \pm 0.1 $ & $ 92.2 \pm 0.2 $ & $ 86.2 \pm 0.7 $ \\
  $\textsc{UB}$ & $ 89.2 \pm 0.5 $ & $ 92.4 \pm 0.4 $ & $ 88.0 \pm 0.1 $ & $ 88.0 \pm 0.5 $ & $ 89.2 \pm 0.5 $ & $ 92.4 \pm 0.2 $ & $ 90.0 \pm 0.4 $ \\ 
  $\Delta_{8,50}^P$ & $ 0.2 \pm 0.1 $ & $ 0.2 \pm 0.2 $ & $ 0.7 \pm 0.2 $ & $ 0.1 \pm 0.1 $ & $ 0.2 \pm 0.1 $ & $ 0.1 \pm 0.0 $ & $ -0.2 \pm 1.7 $ \\
  $\Delta_{8,50}^L$ & $ -0.1 \pm 0.1 $ & $ -0.0 \pm 0.0 $ & $ 0.1 \pm 0.1 $ & $ 0.1 \pm 0.0 $ & $ 0.0 \pm 0.0 $ & $ 0.0 \pm 0.0 $ & $ 0.4 \pm 0.4 $ \\
  % $\Delta_{S}^P$ & $ 0.2 \pm 0.1 $ & $ 0.3 \pm 0.2 $ & $ 0.4 \pm 0.2 $ & $ -0.0 \pm 0.0 $ & $ 0.1 \pm 0.1 $ & $ 0.0 \pm 0.0 $ & $ 0.6 \pm 0.4 $ \\
  % $\Delta_{S}^L$ & $ -0.1 \pm 0.1 $ & $ 0.0 \pm 0.0 $ & $ 0.0 \pm 0.0 $ & $ 0.0 \pm 0.0 $ & $ 0.0 \pm 0.0 $ & $ 0.0 \pm 0.0 $ & $ 0.5 \pm 0.4 $ \\
  $\Delta_{O}^P$ & $ 0.4 \pm 0.1 $ & $ 0.4 \pm 0.2 $ & $ 1.2 \pm 0.3 $ & $ 0.2 \pm 0.0 $ & $ 0.5 \pm 0.0 $ & $ 0.2 \pm 0.0 $ & $ 1.1 \pm 0.6 $ \\ 
  $\Delta_{O}^L$ & $ 0.1 \pm 0.0 $ & $ 0.2 \pm 0.1 $ & $ 0.5 \pm 0.2 $ & $ 0.1 \pm 0.0 $ & $ 0.2 \pm 0.0 $ & $ 0.0 \pm 0.0 $ & $ 0.8 \pm 0.6 $ \\
  \bottomrule
\end{tabular}
\end{adjustbox}
\end{subtable}
\caption{Sentiment analysis results (accuracy).}
\label{tab:res:app:sent}
\end{table*}

\begin{table*}
\small
\begin{subtable}{1\textwidth}
\centering
\begin{adjustbox}{width=\textwidth}
\begin{tabular}{cccccccc} \toprule

  & F $\rightarrow$ G & F $\rightarrow$ SL & F $\rightarrow$ TL & F $\rightarrow$ TR & G $\rightarrow$ F & G $\rightarrow$ SL & G $\rightarrow$ TL \\ 
  \midrule
  \textsc{init} & $ 70.2 \pm 0.8 $ & $ 63.7 \pm 0.8 $ & $ 67.4 \pm 1.3 $ & $ 65.6 \pm 0.8 $ & $ 59.9 \pm 0.8 $ & $ 62.1 \pm 0.5 $ & $ 64.9 \pm 0.9 $ \\ 
  $\textsc{UB}$ & $ 73.8 \pm 0.4 $ & $ 62.6 \pm 0.9 $ & $ 68.3 \pm 0.4 $ & $ 69.9 \pm 0.3 $ & $ 67.6 \pm 0.9 $ & $ 62.6 \pm 0.9 $ & $ 68.3 \pm 0.4 $ \\
  $\Delta_{8,50}^P$ & $ 0.5 \pm 0.5 $ & $ 0.4 \pm 0.4 $ & $ 0.1 \pm 0.4 $ & $ -0.2 \pm 0.4 $ & $ 0.8 \pm 0.2 $ & $ -0.2 \pm 0.2 $ & $ 0.4 \pm 0.3 $ \\
  $\Delta_{8,50}^L$ & $ 0.1 \pm 0.2 $ & $ 0.0 \pm 0.1 $ & $ 0.3 \pm 0.2 $ & $ 0.1 \pm 0.1 $ & $ 0.7 \pm 0.4 $ & $ -0.2 \pm 0.1 $ & $ 0.1 \pm 0.1 $ \\ 
  % $\Delta_{S}^P$ & $ 0.4 \pm 0.4 $ & $ 0.2 \pm 0.4 $ & $ 0.1 \pm 0.4 $ & $ -0.0 \pm 0.3 $ & $ 1.0 \pm 0.7 $ & $ -0.6 \pm 0.4 $ & $ -0.5 \pm 0.6 $ \\
  % $\Delta_{S}^L$ & $ 0.3 \pm 0.2 $ & $ 0.1 \pm 0.2 $ & $ 0.5 \pm 0.3 $ & $ 0.1 \pm 0.1 $ & $ 0.9 \pm 0.5 $ & $ -0.1 \pm 0.0 $ & $ -0.1 \pm 0.1 $ \\
  $\Delta_{O}^P$ & $ 1.2 \pm 0.4 $ & $ 0.9 \pm 0.3 $ & $ 0.9 \pm 0.3 $ & $ 0.7 \pm 0.2 $ & $ 1.8 \pm 0.6 $ & $ 0.4 \pm 0.1 $ & $ 1.2 \pm 0.2 $ \\ 
  $\Delta_{O}^L$ & $ 0.6 \pm 0.2 $ & $ 0.6 \pm 0.2 $ & $ 0.8 \pm 0.3 $ & $ 0.5 \pm 0.2 $ & $ 1.5 \pm 0.5 $ & $ 0.2 \pm 0.0 $ & $ 0.9 \pm 0.2 $ \\
%   \bottomrule
\end{tabular}
\end{adjustbox}
\end{subtable}
\begin{subtable}{1\textwidth}
\centering
\begin{adjustbox}{width=\textwidth}
\begin{tabular}{cccccccc} \midrule

  & G $\rightarrow$ TR & SL $\rightarrow$ F & SL $\rightarrow$ G & SL $\rightarrow$ TL & SL $\rightarrow$ TR & TL $\rightarrow$ F & TL $\rightarrow$ G \\ 
  \midrule
  \textsc{init} & $ 68.8 \pm 0.2 $ & $ 62.0 \pm 1.6 $ & $ 71.1 \pm 1.4 $ & $ 63.7 \pm 1.2 $ & $ 67.0 \pm 1.2 $ & $ 63.6 \pm 0.5 $ & $ 69.7 \pm 0.4 $ \\
  $\textsc{UB}$ & $ 69.9 \pm 0.3 $ & $ 67.6 \pm 0.9 $ & $ 73.8 \pm 0.4 $ & $ 68.3 \pm 0.4 $ & $ 69.9 \pm 0.3 $ & $ 67.6 \pm 0.9 $ & $ 73.8 \pm 0.4 $ \\ 
  $\Delta_{8,50}^P$ & $ -0.0 \pm 0.1 $ & $ 0.8 \pm 0.4 $ & $ -0.5 \pm 0.2 $ & $ 1.1 \pm 0.4 $ & $ -0.1 \pm 0.1 $ & $ -0.6 \pm 0.3 $ & $ -1.1 \pm 0.6 $ \\
  $\Delta_{8,50}^L$ & $ -0.1 \pm 0.1 $ & $ 0.4 \pm 0.2 $ & $ 0.1 \pm 0.1 $ & $ 0.7 \pm 0.1 $ & $ 0.1 \pm 0.2 $ & $ 0.2 \pm 0.1 $ & $ -0.2 \pm 0.1 $ \\ 
  % $\Delta_{S}^P$ & $ 0.1 \pm 0.0 $ & $ 0.7 \pm 0.3 $ & $ -0.3 \pm 0.2 $ & $ 1.4 \pm 0.5 $ & $ -0.2 \pm 0.1 $ & $ -0.7 \pm 0.6 $ & $ 0.0 \pm 0.1 $ \\
  % $\Delta_{S}^L$ & $ -0.2 \pm 0.0 $ & $ 1.1 \pm 0.4 $ & $ -0.2 \pm 0.1 $ & $ 1.1 \pm 0.2 $ & $ 0.1 \pm 0.2 $ & $ 0.1 \pm 0.1 $ & $ -0.1 \pm 0.0 $ \\
  $\Delta_{O}^P$ & $ 0.5 \pm 0.1 $ & $ 1.5 \pm 0.4 $ & $ 0.3 \pm 0.1 $ & $ 2.0 \pm 0.5 $ & $ 0.5 \pm 0.1 $ & $ 0.7 \pm 0.2 $ & $ 0.7 \pm 0.2 $ \\
  $\Delta_{O}^L$ & $ 0.2 \pm 0.1 $ & $ 1.4 \pm 0.4 $ & $ 0.3 \pm 0.1 $ & $ 1.4 \pm 0.2 $ & $ 0.6 \pm 0.1 $ & $ 0.6 \pm 0.1 $ & $ 0.3 \pm 0.0 $ \\

%   \bottomrule
\end{tabular}
\end{adjustbox}
\end{subtable}
\begin{subtable}{1\textwidth}
\centering
\begin{adjustbox}{width=\textwidth}
\begin{tabular}{ccccccc|c} \midrule

  & TL $\rightarrow$ SL & TL $\rightarrow$ TR & TR $\rightarrow$ F & TR $\rightarrow$ G & TR $\rightarrow$ SL & TR $\rightarrow$ TL & AVG \\ 
  \midrule
  \textsc{init} & $ 61.6 \pm 0.5 $ & $ 64.9 \pm 0.5 $ & $ 60.0 \pm 1.0 $ & $ 71.5 \pm 0.7 $ & $ 61.3 \pm 0.6 $ & $ 63.3 \pm 1.1 $ & $ 65.1 \pm 0.9 $ \\ 
  $\textsc{UB}$ & $ 62.6 \pm 0.9 $ & $ 69.9 \pm 0.3 $ & $ 67.6 \pm 0.9 $ & $ 73.8 \pm 0.4 $ & $ 62.6 \pm 0.9 $ & $ 68.3 \pm 0.4 $ & $ 68.4 \pm 0.7 $ \\ 
  $\Delta_{8,50}^P$ & $ -0.3 \pm 0.4 $ & $ -0.5 \pm 0.4 $ & $ -0.1 \pm 0.5 $ & $ -0.1 \pm 0.2 $ & $ 0.1 \pm 0.2 $ & $ 0.4 \pm 0.3 $ & $ 0.0 \pm 0.4 $ \\
  $\Delta_{8,50}^L$ & $ 0.5 \pm 0.2 $ & $ -0.4 \pm 0.3 $ & $ 0.3 \pm 0.5 $ & $ 0.3 \pm 0.3 $ & $ 0.0 \pm 0.1 $ & $ 0.3 \pm 0.3 $ & $ 0.2 \pm 0.2 $ \\ 
  % $\Delta_{S}^P$ & $ -2.2 \pm 0.7 $ & $ 0.1 \pm 0.1 $ & $ 0.8 \pm 0.4 $ & $ -0.0 \pm 0.2 $ & $ 0.2 \pm 0.2 $ & $ 0.0 \pm 0.1 $ & $ 0.0 \pm 0.3 $ \\
  % $\Delta_{S}^L$ & $ 0.3 \pm 0.1 $ & $ 0.2 \pm 0.1 $ & $ 0.4 \pm 0.6 $ & $ 0.1 \pm 0.2 $ & $ 0.2 \pm 0.2 $ & $ 0.1 \pm 0.1 $ & $ 0.2 \pm 0.2 $ \\
  $\Delta_{O}^P$ & $ 1.2 \pm 0.1 $ & $ 0.7 \pm 0.1 $ & $ 1.7 \pm 0.4 $ & $ 0.7 \pm 0.2 $ & $ 0.8 \pm 0.2 $ & $ 1.2 \pm 0.3 $ & $ 1.0 \pm 0.3 $ \\ 
  $\Delta_{O}^L$ & $ 1.1 \pm 0.2 $ & $ 0.6 \pm 0.1 $ & $ 1.0 \pm 0.4 $ & $ 0.7 \pm 0.2 $ & $ 0.6 \pm 0.2 $ & $ 0.8 \pm 0.3 $ & $ 0.7 \pm 0.2 $ \\

  \bottomrule
\end{tabular}
\end{adjustbox}
\end{subtable}
\caption{MNLI results (macro-F1).}
\label{tab:res:app:mnli}
\end{table*}

\begin{table*}
\small
\begin{subtable}{1\textwidth}
\centering
\begin{adjustbox}{width=\textwidth}
\begin{tabular}{cccccccc} \toprule
& D $\rightarrow$ L & D $\rightarrow$ R & D $\rightarrow$ S & L $\rightarrow$ D & L $\rightarrow$ R & L $\rightarrow$ S & R $\rightarrow$ D \\ 
\midrule
\textsc{init} & $ 50.9 \pm 0.8 $ & $ 36.9 \pm 1.1 $ & $ 40.5 \pm 0.9 $ & $ 47.6 \pm 0.2 $ & $ 35.3 \pm 0.8 $ & $ 36.3 \pm 0.5 $ & $ 46.2 \pm 0.9 $ \\
$\textsc{UB}$ & $ 85.5 \pm 0.3 $ & $ 83.4 \pm 0.2 $ & $ 81.2 \pm 0.2 $ & $ 67.1 \pm 0.5 $ & $ 83.4 \pm 0.2 $ & $ 81.2 \pm 0.2 $ & $ 67.1 \pm 0.5 $ \\ 
$\Delta_{8,50}^P$ & $ -1.2 \pm 0.6 $ & $ -3.0 \pm 1.2 $ & $ -2.2 \pm 1.0 $ & $ 0.9 \pm 0.1 $ & $ 3.6 \pm 0.7 $ & $ 1.0 \pm 0.3 $ & $ 1.3 \pm 0.3 $ \\
$\Delta_{8,50}^L$ & $ -0.2 \pm 0.1 $ & $ -0.4 \pm 0.3 $ & $ -0.1 \pm 0.2 $ & $ 0.2 \pm 0.1 $ & $ 0.3 \pm 0.2 $ & $ 0.2 \pm 0.1 $ & $ 0.1 \pm 0.1 $ \\
% $\Delta_{S}^P$ & $ -0.1 \pm 0.1 $ & $ 0.0 \pm 0.0 $ & $ -0.0 \pm 0.2 $ & $ 0.5 \pm 0.1 $ & $ 4.1 \pm 1.2 $ & $ 0.4 \pm 0.5 $ & $ 0.2 \pm 0.1 $ \\
% $\Delta_{S}^L$ & $ -0.1 \pm 0.1 $ & $ 0.1 \pm 0.1 $ & $ 0.1 \pm 0.0 $ & $ 0.2 \pm 0.1 $ & $ 0.5 \pm 0.1 $ & $ 0.0 \pm 0.0 $ & $ 0.0 \pm 0.2 $ \\
$\Delta_{O}^P$ & $ 0.3 \pm 0.2 $ & $ 0.6 \pm 0.3 $ & $ 0.2 \pm 0.2 $ & $ 1.4 \pm 0.1 $ & $ 6.7 \pm 1.0 $ & $ 1.9 \pm 0.4 $ & $ 2.1 \pm 0.5 $ \\
$\Delta_{O}^L$ & $ 0.2 \pm 0.1 $ & $ 0.3 \pm 0.2 $ & $ 0.4 \pm 0.1 $ & $ 0.7 \pm 0.0 $ & $ 1.5 \pm 0.3 $ & $ 0.5 \pm 0.2 $ & $ 0.7 \pm 0.2 $ \\
%   \bottomrule
\end{tabular}
\end{adjustbox}
\end{subtable}
\begin{subtable}{1\textwidth}
\centering
\begin{adjustbox}{width=\textwidth}
\begin{tabular}{cccccc|c} \midrule
 & R $\rightarrow$ L & R $\rightarrow$ S & S $\rightarrow$ D & S $\rightarrow$ L & S $\rightarrow$ R & AVG \\ 
 \midrule
  \textsc{init} & $ 44.1 \pm 1.1 $ & $ 33.2 \pm 0.9 $ & $ 49.1 \pm 0.3 $ & $ 44.9 \pm 0.5 $ & $ 55.6 \pm 0.6 $ & $ 43.4 \pm 0.8 $ \\ 
  $\textsc{UB}$ & $ 85.5 \pm 0.3 $ & $ 81.2 \pm 0.2 $ & $ 67.1 \pm 0.5 $ & $ 85.5 \pm 0.3 $ & $ 83.4 \pm 0.2 $ & $ 79.3 \pm 0.4 $ \\ 
  $\Delta_{8,50}^P$ & $ 9.5 \pm 0.8 $ & $ 11.2 \pm 0.7 $ & $ 0.6 \pm 0.2 $ & $ -2.1 \pm 0.4 $ & $ -4.2 \pm 0.7 $ & $ 1.3 \pm 0.7 $ \\ 
  $\Delta_{8,50}^L$ & $ 2.2 \pm 0.5 $ & $ 2.4 \pm 0.6 $ & $ 0.0 \pm 0.1 $ & $ -0.5 \pm 0.2 $ & $ -0.7 \pm 0.4 $ & $ 0.3 \pm 0.3 $ \\ 
  % $\Delta_{S}^P$ & $ 12.1 \pm 1.6 $ & $ 18.0 \pm 0.8 $ & $ 0.5 \pm 0.2 $ & $ -0.4 \pm 0.3 $ & $ 0.1 \pm 0.1 $ & $ 2.9 \pm 2.5 $ \\ 
  % $\Delta_{S}^L$ & $ 4.1 \pm 1.0 $ & $ 6.2 \pm 0.9 $ & $ 0.0 \pm 0.0 $ & $ -0.1 \pm 0.2 $ & $ 0.1 \pm 0.0 $ & $ 0.9 \pm 0.8 $ \\
  $\Delta_{O}^P$ & $ 14.4 \pm 0.9 $ & $ 18.8 \pm 0.9 $ & $ 0.9 \pm 0.2 $ & $ 0.3 \pm 0.2 $ & $ 0.3 \pm 0.2 $ & $ 4.0 \pm 0.5 $ \\ 
  $\Delta_{O}^L$ & $ 5.7 \pm 0.9 $ & $ 6.8 \pm 0.7 $ & $ 0.3 \pm 0.1 $ & $ 0.2 \pm 0.1 $ & $ 0.2 \pm 0.1 $ & $ 1.5 \pm 0.4 $ \\

  \bottomrule
\end{tabular}
\end{adjustbox}
\end{subtable}
\caption{Aspect prediction results (binary-F1).}
\label{tab:res:app:aspect}
\end{table*}

\section{Performance for different $\beta$}
\label{appendix:hyper}
\begin{figure}
    \centering
    \includegraphics[width=1\columnwidth]{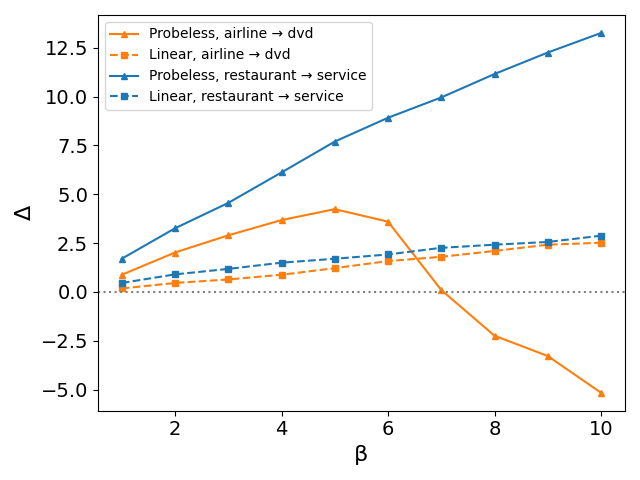}
    \caption{Results for different $\beta$ values, using $k=50$.}
    \label{fig:beta}
\end{figure}
While our default hyperparameter values, $\beta=8$ and $k=50$ improve performance in most cases, they are not optimal for all cases.
Fig.~\ref{fig:beta} shows that when $k=50$, the optimal $\beta$ value for the Airline $\rightarrow$ DVD case is $5$, whereas for Restaurants $\rightarrow$ Service it is actually better to use a greater $\beta$.
Thus, it is not possible to find one value that would be optimal for all cases.

\end{document}